\documentclass[final]{l4dc2024}

\title[Real-Time Safe Control of NNDMs]{Real-Time Safe Control of Neural Network Dynamic Models\\ with Sound Approximation}

\usepackage{xcolor}

\newcommand{\UU}{\mathcal{U}}

\newcommand{\XX}{\mathcal{X}}

\newcommand{\real}{\mathbb{R}}

\newcommand{\ie}{\textit{i.e. }}
\newcommand{\st}{\textit{s.t. }}
\newcommand{\eg}{\textit{e.g. }}

\DeclareMathOperator*{\argmax}{arg\,max}

\renewcommand{\v}[1]{\boldsymbol{\mathbf{#1}}}

\newcommand{\cor}[1]{{\color{orange} #1}}
\newcommand{\cbl}[1]{{\color{blue} #1}}

\newcommand{\cgr}[1]{{\color{teal} #1}}
\newcommand{\cbr}[1]{{\color{brown} #1}}
\newtheorem{assumption}{Assumption}
\usepackage[normalem]{ulem}
\usepackage{times}
\usepackage{cleveref}
\usepackage{multicol}
\usepackage{multirow}
\usepackage{booktabs}
\usepackage{graphicx}
\usepackage{subcaption}
\usepackage[font=footnotesize]{caption}
\usepackage{subcaption}
\usepackage{caption}
\usepackage{enumitem}
\usepackage{wrapfig}
\usepackage{comment}

\author{%
 \Name{Hanjiang Hu} \Email{hanjianghu@cmu.edu}\\
 \addr Robotics Institute, Carnegie Mellon University
 \AND
 \Name{Jianglin Lan} \Email{jianglin.lan@glasgow.ac.uk}\\
 \addr James Watt School of Engineering, University of Glasgow
  \AND
 \Name{Changliu Liu} \Email{cliu6@andrew.cmu.edu}\\
 \addr Robotics Institute, Carnegie Mellon University%
}

\begin{document}

\maketitle

\begin{abstract}%
Safe control of neural network dynamic models (NNDMs) is important to robotics and many applications. However, it remains challenging to compute an optimal safe control in real time for NNDM. To enable real-time computation, we propose to use a sound approximation of the NNDM in the control synthesis. In particular, we propose Bernstein over-approximated neural dynamics (BOND) based on the Bernstein polynomial over-approximation (BPO) of ReLU activation functions in NNDM. 
To mitigate the errors introduced by the approximation and to ensure persistent feasibility of the safe control problems, we synthesize a worst-case safety index using the most unsafe approximated state within the BPO relaxation of NNDM offline. For the online real-time optimization, we formulate the first-order Taylor approximation of the nonlinear worst-case safety constraint as an additional linear layer of NNDM with the $\ell_2$ bounded bias term for the higher-order remainder. Comprehensive experiments with different neural dynamics and safety constraints show that with safety guaranteed, our NNDMs with sound approximation are 10-100 times faster than the safe control baseline that uses mixed integer programming (MIP), validating the effectiveness of the worst-case safety index and scalability of the proposed BOND in real-time large-scale settings.

\end{abstract}

\begin{keywords}%
safe control, neural network dynamic model, Bernstein polynomial
\end{keywords}

\section{Introduction}

Safety is crucial to robotic systems. Safe control of dynamic systems has been well studied in literature \citep{nagumo1942lage, blanchini1999set}. A safe control law maintains the states within the user-defined safety set by ensuring forward invariance and finite-time convergence, \ie states remaining in it once entering; and returning to it in finite time steps once leaving. Although safe control laws can be designed for control-affine systems \citep{ liu2014control,wei2019safe,agrawal2021safe}, it is challenging to construct an exact analytical model for complex real-world systems. Progress in deep neural networks has boosted learning-based methods to model the complicated dynamics \citep{nagabandi2018neural, janner2019trust}. However, these neural network dynamic models (NNDMs) have limited mathematical interpretability, making it  difficult to design subsequent safe control laws.

This paper considers the safe tracking problem using NNDMs. Recent works \citep{wei2022safe, liu2023model,li2023system} model the safe tracking problem as a constrained optimization problem by minimizing the state tracking error for the given system dynamics constraint while obeying the safety constraint. Since it is challenging to find optimal tracking control for these highly nonlinear black-box NNDMs through model inverse \citep{tolani2000real} or shooting methods, researchers resort to mixed integer programming (MIP) to find the optimal tracking control \citep{wei2022safe, liu2023model, li2023system}, a method widely used in neural network verification. However, MIP is well-known for its poor time efficiency and limited scalability in the literature on neural network verification \citep{liu2021algorithms, li2023sok}, making these complete MIP-based methods hardly applicable in real-time safety-critical robot applications.

To this end, we propose Bernstein over-approximated neural dynamics (BOND) with  sound approximation of ReLU activation layers in NNDM to greatly speed up the computation of safe tracking problems. Specifically, inspired by sound verification of neural networks \citep{fatnassi2023bern,huang2022polar, khedr2023deepbern}, we leverage Bernstein polynomial over-approximation (BPO) to address the nonlinearity of the activation function,  replacing integer variables with inequality constraints for NNDMs in the safe tracking optimization. To deal with the approximation error caused by BPO and ensure persistent feasibility under safety constraints, we synthesize the worst-case safety index offline to make the optimization problem feasible even for the most potentially unsafe  state of BOND, and linearize the safety constraint with a linear Taylor layer in the online optimization.  The contributions are listed as below:
\begin{itemize}[leftmargin=0.4cm]
    \item We propose a sound approximation for NNDM using Bernstein polynomial over-approximation to optimize real-time safe tracking problems efficiently.
    \item We synthesize the worst-case safety index to ensure the persistent feasibility under approximation error caused by the over-approximation of NNDMs.
    \item Extensive experiments validate that BOND is 10-100 times faster and more scalable than MIP-based baseline in  real-time collision avoidance and safe following with different NNDMs.
\end{itemize}
The remaining paper is organized as follows: \Cref{sec:formulation} provides a problem formulation of neural network dynamics, Bernstein polynomial over-approximation and safe tracking problem. \Cref{sec:method} describes the proposed method including worst-case safety index synthesis and linearization for online optimization. \Cref{sec:experiment} presents the experimental results with ablation study. \Cref{sec:conclusion} concludes the paper and discusses potential future directions. 




\section{Formulation}
\label{sec:formulation}

\subsection{Background of Safe Tracking with Neural Network Dynamic Models}
\label{sec:tracking_NNDM}
\paragraph{Neural network dynamic models (NNDMs).}Denote a discrete-time NNDM with state $\v x_k$ and control $\v u_k$ at time step $k$ as 
\begin{align}
\label{eq:nndm}
\v x_{k+1} = \v x_k + \v f(\v x_k, \v u_k) dt,~~ \v x_k  \in \XX = \real^{m_x},~~\v u_k\in \UU \subset \real^{m_u}
\end{align}
where $\UU$ is defined by linear constraints and $dt$ is the sampling time. $\v f: \real^{m_x} \times \real^{m_u} \mapsto \real^{m_x}$ is the dynamic model parameterized by $n$-layer feedforward neural networks with nonlinear activation functions, $\ie \v{f} = \v{f}_n \circ \v{f}_{n-1} \circ \cdots \circ \v{f}_1$, where $\v{f}_i: \real^{k_{i-1}}\mapsto \real^{k_i}$ is the $i$th linear mapping layer with a nonlinear activation $\v \sigma_i: \real^{k_{i}}\mapsto \real^{k_i}$ over the $k_i$-dimensional hidden variable in layer $i$. More concretely, by denoting the weight matrix and bias vector in layer $i$ as $\v W_i \in \real^{k_i\times k_{i-1}}$ and $\v b_i \in \real^{k_i}$, the hidden variable after the layer $i$ is $\v z_i = \v f_i(\v z_{i-1})=\v \sigma_i(\hat{\v z}_i)$, where $\hat{\v z}_i = \v W_i \v z_{i-1} + \v b_i$ is the pre-activation variable. Specifically, it trivially holds that $\v z_0 = [\v x_k^\top,\v u_k^\top]^\top,k_0=m_x+m_u,k_n=m_x$. Let $\v w_{ij}\in\real^{1\times k_{i-1}}$ be the $j$th row of $\v W_i$ and $b_{ij}$ be the $j$th entry of $\v b_i$, so the $j$th entry of $\v{\hat z}_i$ is calculated as $\hat z_{ij}=\v w_{ij} \v z_{i-1} + b_{i,j}$. We only focus on ReLU activation in this work, so for the $j$th entry of $\v{z}_i$, we have that $z_{ij} = \sigma_i(\hat{z}_{ij}) = \max\{0, \hat{z}_{ij}\}$. 

\paragraph{Optimization problem for tracking.} Similar to \cite{wei2022safe}, we focus on the tracking problem with the NNDM as a one-step model predictive control (MPC), optimizing control action $\v u_k$ via minimizing the $\ell_p$ distance between the predicted next state $\v x_{k+1}$ and the reference next state $\v x_{k+1}^r$ (known ahead of time) at each time step $k$, which is shown as follows:
\begin{align}
    \label{eq:opt_no_safe}
    \begin{split} 
         & \min_{\v u_k, \v x_{k+1}}  \|\v x_{k+1} - \v x_{k+1}^r\|_p\\ 
        \st & \v x_{k+1} = \v x_k + \v f(\v x_k, \v u_k) dt,\quad \v u_k\in \UU.\\
    \end{split}
\end{align}
where $\|\cdot\|_p$ can be either $\ell_1$-norm as a linear objective or $\ell_2$-norm as a quadratic objective. Under the nonlinear constraint of NNDM with ReLU activation, the optimization problem \eqref{eq:opt_no_safe} is challenging to solve using existing solvers \citep{wei2022safe, liu2023model}.

\paragraph{Safety specification and constraint.} In addition to the NNDM constraint in \eqref{eq:opt_no_safe}, the safety constraint is also indispensable for the safe tracking problem \citep{wei2022safe}. 
We denote the user-specified safe set $\XX_0$ as a connected and closed set in the state space, which can be defined as a zero-sublevel set of a continuous and differentiable function, \ie $\XX_0=\{\v x\in\XX \mid \phi_0(\v x)\leq 0\}$. If the system is already in a safe state, we should ensure forward invariance, \ie $  \phi_0(\v x_{k}) \leq 0 \implies \phi_0(\v x_{k+1}) \leq 0$. If the system is currently unsafe, we should ensure finite-time convergence, \ie  $\phi_0(\v x_k) > 0 \implies \phi_0(\v x_{k+1}) \leq \phi_0(\v x_{k}) - \gamma dt$, so that the system will go back to the safe set within finite time steps $\phi_0(\v x_k)/\gamma dt$ with constant $\gamma$. We combine these two constraints at step $k$ as follows,
\begin{align}
    \mathcal{A}(\XX_0, \v x_{k},\gamma):=\{\v x_{k+1}\mid \phi_0(\v x_{k+1})\leq\max\{0,\phi_0(\v x_k)-\gamma dt\}, \text{ with } \XX_0=\{\v x \mid \phi_0(\v x)\leq 0\}.\label{eq:k-step-safe-set}
\end{align}
 However, there may not always exist a feasible control input that results in $\v x_{k+1}\in \mathcal{A}(\XX_0, \v x_{k},\gamma)$. If such control always exists, we say the safe tracking problem is \textit{persistently feasible}. To achieve persistent feasibility, the common practice is to find a subset of the safe set $\XX_S\subset \XX_0$ using safety index synthesis (SIS) \citep{wei2022safe} such that there always exist a control that ensures $\v x_{k+1}\in \mathcal{A}(\XX_S, \v x_{k},\gamma)$. Then $\mathcal{A}(\XX_S, \v x_{k},\gamma)$ will be used as a constraint in \eqref{eq:opt_no_safe} to ensure safety. 

\begin{figure}
    \centering
    \subfigure[]{\includegraphics[width=0.5\textwidth]{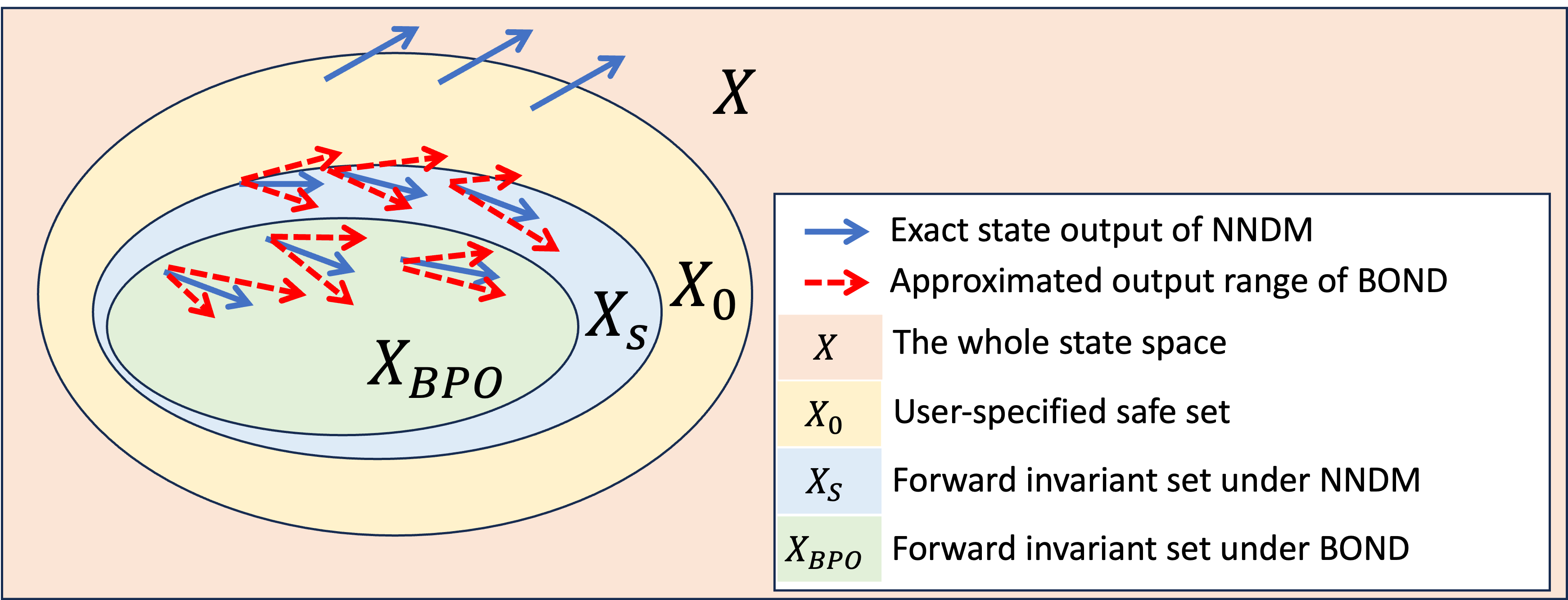}
    } 
    \subfigure[]{\includegraphics[width=0.22\textwidth]{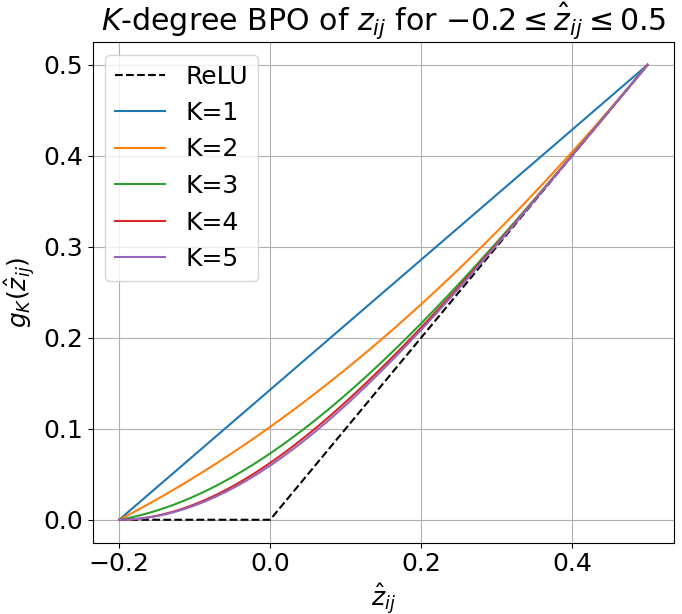}} 
    \vspace{-4mm}
    \caption{(a) Illustration of the forward invariant sets under exact NNDM and BOND. (b) Approximation of the ReLU function using BPO of different degrees.}
    \vspace{-5mm}
    \label{fig:overview}
\end{figure}

\subsection{Safe Tracking with Bernstein Over-approximated Neural Dynamics (BOND)}
\label{sec:BPO}
\paragraph{Bernstein polynomial over-approximation for NNDMs.} Inspired by neural network verification with Bernstein polynomial over-approximation (BPO) \citep{fatnassi2023bern,huang2022polar, khedr2023deepbern}, we adopt the following $K$-order BPO for  the ReLU activation $z_{ij} = \sigma_i(\hat{z}_{ij})$ based on the bounded pre-activation values $\hat l_{ij} \leq \hat z_{ij} \leq \hat u_{ij}, \hat{u}_{ij} \neq \hat{l}_{ij}$:
\begin{align}
\label{eq:BPO}
 z_{ij}\leq g_K(\hat z_{ij}) = \sum_{k=0}^K \max\{0, \frac{k}{K}(\hat u_{ij}-\hat l_{ij}) + \hat l_{ij}\}\cdot\tbinom{K}{k}\frac{(\hat u_{ij}-\hat z_{ij})^{K-k}(\hat z_{ij}-\hat l_{ij})^k}{(\hat u_{ij}-\hat l_{ij})^K},
\end{align}
where the pre-activation bounds $\hat l_{ij}$ and $\hat u_{ij}$ can be found through interval arithmetic (IA) methods or dual networks \citep{liu2021algorithms, wong2018provable}. When $\hat{u}_{ij} = \hat{l}_{ij}$, it trivially holds that $z_{ij}=\sigma_i (\hat z_{i,j})=\sigma_i (\hat l_{i,j})=\sigma_i (\hat u_{i,j})$ so the approximation is not needed.
 Note that when $K=1$,  BPO relaxation is degraded to the triangle relaxation (also called LP relaxation) \citep{wong2018scaling, tjeng2018evaluating, ehlers2017formal}. The nonlinear higher-degree BPOs are visualized in Fig. \ref{fig:overview} (b).

\paragraph{Conservative forward invariant set with BPO.} Combining BPO \eqref{eq:BPO} with the linear under approximation for ReLU activation at each node, $ z_{ij}\geq 0, z_{ij}\geq \hat z_{ij}$, the output of the dynamic model $\v f(\v x, \v u)$ can be lower-bounded by a function $\v{\underline{f}}_B$ and upper-bounded by a function $\v{\bar{f}}_B$, \ie $\v{\underline f}_{B}(\v x, \v u)\leq \v f(\v x, \v u)\leq\v{\bar f}_{B}(\v x, \v u)$, relaxing the predicted output from the exact $\v f(\v x, \v u)$ to the range $[\v{\underline f}_{B}(\v x, \v u), \v{\bar f}_{B}(\v x, \v u)]$. Therefore, to ensure persistent feasibility of the safety constraints w.r.t the unknown output $\v f(\v x, \v u)\in [\v{\underline f}_{B}(\v x, \v u), \v{\bar f}_{B}(\v x, \v u)]$,
we need to find an even more conservative forward invariant set $\XX_{BPO}\subseteq \XX_S$, which is illustrated in Fig. \ref{fig:overview} (a) and will be introduced in \cref{sec:method}.

\paragraph{Safe tracking with BOND.}
 With the BPO relaxation of NNDM and more conservative forward invariant set $\XX_{BPO}$, 
 the tracking problem aims to optimize both the control strategy $\v u_k$ and the hallucinated next state $\v x_{k+1}$ within the approximated output range $[\v{\underline f}_{B}(\v x, \v u), \v{\bar f}_{B}(\v x, \v u)]$ by minimizing a linear ($p=1$) or quadratic ($p=2$) objective of the distance between  $\v x_{k+1}$ and the reference state $\v x_{k+1}^r$ whilst satisfying the safe control constraint that any possible future state should belong to $\XX_{BPO}$. 
So the safe tracking problem with BOND $\v f_B(\cdot)$ is formulated as the  following constrained optimization problem at every time step $k$: 
\begin{subequations}\label{eq: bond opt}
\begin{align}
    \label{eq:safe_control_minimax}
 &\min_{\v u_k\in\UU,\v x_{k+1}}  \|\v x_{k+1} - \v x_{k+1}^r\|_p 
\\ \label{eq:BPO_NNDM}
\st & \v x_k + \v{\underline f}_B(\v x_k, \v u_k) dt \leq \v x_{k+1} \leq \v x_k + \v{\bar f}_B(\v x_k, \v u_k) dt\\ \label{eq:BPO_safe}
 & \v{\tilde x}_{k+1}\in\mathcal{A}(\XX_{BPO}, \v x_{k},\gamma), \forall \v{\tilde x}_{k+1}\in[\v x_k + \v{\underline f}_B(\v x_k, \v u_k) dt , \v x_k + \v{\bar f}_B(\v x_k, \v u_k) dt].
\end{align}
\end{subequations}
Note that $\XX_S$ from \cite{wei2022safe} will not work in \eqref{eq:BPO_safe} as it does not take the approximation into consideration.  In the following section, we first discuss how to obtain $\XX_{BPO}$ offline; and then discuss how to efficiently compute the optimization problem \eqref{eq: bond opt} online.



\section{Method}
\label{sec:method}

\subsection{Worst-Case Safety Index Synthesis}
\label{sec:worstcase_SI}
To characterize the forward invariant set $\XX_{BPO}$ and $\mathcal{A}(\XX_{BPO}, \v x_{k},\gamma)$, similar to \eqref{eq:k-step-safe-set}, we propose to synthesize a worst-case safety index $\phi$ so that $\XX_{BPO}=\{\v x\in\XX \mid \phi(\v x)\leq 0\}$ and \eqref{eq:BPO_safe} is persistently feasible. 
Following the evolutionary strategy-based safety index synthesis in \cite{wei2022safe}, we parameterize the safety index  with $\alpha_i,\beta\in\real, i=1,2,\dots,q$ as 
$\phi(\v x) = \phi_0^*(\v \alpha_0, \v x) + \sum_{i=1}^q \alpha_i \phi_0^{(i)}(\v x) + \beta$, where  $\phi_0^*(\v \alpha_0, \v x)$ is consistent with user-specified sublevel set $\phi_0$ and $\phi_0^{(i)}(\v x)$ is denoted as each $i$the order derivative of $\phi_0$ to ensure the relative degree of 1 from $\phi_0^{(q)}$ to $\v u$~\citep{liu2014control}. 
Therefore, the safety constraint in \eqref{eq:BPO_safe} can be equivalently written as 
\begin{align}
    \phi(\v x_{k+1}^{wc}) \leq \max\{0, \phi(\v x_{k}) - \gamma dt\}, 
\label{eq:worst_case_safety}
\end{align}
where $\v x_{k+1}^{wc}=\v x_k + \v f^{wc}(\v x_k,\v u_k) dt$  and $\v f^{wc}(\v x_k,\v u_k) = \argmax_{ \v{\underline f}_{B}(\v x_k, \v u_k) \leq \v f\leq \v{\bar f}_{B}(\v x_k, \v u_k)}\phi(\v x_k+ \v f dt)$ is the worst case state and the worst case NN relaxation, respectively. Our goal is to find $\phi$ such that for all state $\v x_k$, there exists a control $\v u_k\in\mathcal{U}$ that satisfies \eqref{eq:worst_case_safety} (persistent feasibility). 
For the BPO-relaxed dynamics, we define the whole legal state set without safety constraint as $B\subseteq \XX$ and 
the infeasible state subset of $B$ regarding persistent feasibility as $B^*=\{\v x_k\in B\mid \forall \v u_k, \phi(\v x_{k+1}^{wc}) > \max\{0, \phi(\v x_{k}) - \gamma dt\}\}$. The emptiness of $  B^*=\emptyset$ implies persistent feasibility. 

Following \cite{wei2022safe}, we adopt the implementation \citep{Feldt2018} of  evolutionary methods  \citep{das2016recent,hansen2016cma} to optimize the parameters in $\phi$. Specifically, the evolution algorithm runs for multiple generations. In each iteration,  we uniformly sample a dense subset $S \subset B$ and find the minimal infeasible rate $r = |S\cap B^*|/|S|$ based on the sampled parameter candidates from a multivariate Gaussian distribution. The new Gaussian distribution will be updated based on the last candidates with the least infeasible rate $r$.  Besides, through reachability-based methods like interval arithmetic, the Euclidean error of $\v f$ can be upper-bounded by $\Delta f= \max_{\v x, \v u}\|\v f_B - \v f\|$.
Therefore, we propose the following Proposition \ref{prop: feasibility} based on Assumption \ref{asp: lipschitz}, showing that with dense sampling $S\in B$ and the convergence of $r$ to 0,  the optimized safety index can induce persistent feasibility even with the worst-case unsafe state update $\v f^{wc}(\v x,\v u)$ for any state in $B$.
\begin{assumption}\label{asp: lipschitz}
$\v f$ and $\phi$ are Lipschitz continuous functions over compact set $B\subseteq \mathcal{X}$ with Lipschitz constants  $k_f$ and $k_\phi$ under $\ell_2$ norm, respectively. 
\end{assumption}

\begin{proposition}\label{prop: feasibility}
Suppose 1) we sample a state subset $S \subset B$ such that  $\forall \v x \in B$, $\min_{\v x' \in S } \|\v x - \v x'\| \leq \delta$, where $\delta$ is the sampling density; and 2) $\forall \v x' \in S$, there exists a safe control $\v u$, \st $\phi(\v x' + \v f^{wc}(\v x',\v u)dt) \leq \max\{-\epsilon, \phi(\v x') - \gamma dt -\epsilon\}$,
where $\epsilon = k_{\phi} (2\delta + 2\Delta f dt+k_f\delta dt)$. Then $\forall \v x \in B, \exists \v u,\ \st$
\begin{align}\label{eq: lemma 1 inequality}
\phi(\v x^{wc})=    \max_{\v f(\v x, \v u)\in[ \v{\underline f}_{B}(\v x, \v u) ,\v{\bar f}_{B}(\v x, \v u) ]}\phi(\v x + \v f(\v x,\v u)dt) \leq \max\{0, \phi(\v x ) - \gamma dt\}.
\end{align}
\end{proposition}

\begin{proof}
Based on 1), $\forall \v x\in B$, we can find $\v x'\in S$ such that $\|\v x-\v x'\|\leq \delta$. Based on 2), for this $\v x'$, we can find $\v u$ such that $\phi(\v x' + \v f^{wc}(\v x',\v u)dt) \leq \max\{0,\phi(\v x')-\gamma dt\}-\epsilon$. Based on Assumption \ref{asp: lipschitz}, we show below that Eq. \eqref{eq: lemma 1 inequality} holds by using the \cor{Lipschitz condition} $k_f$ and $k_\phi$ and \underline{triangle inequality}:
\begin{align}
     \phi(\v x^{wc})  = &\cor{\phi(\v x + \v f^{wc}(\v x,\v u)dt) - \phi(\v x' + \v f^{wc}(\v x',\v u)dt)} \cgr{ +\phi(\v x' + \v f^{wc}(\v x',\v u)dt)} \nonumber\\
    \leq &\underline{\cor{k_{\phi}\|\v x - \v x' + [\v f^{wc}(\v x,\v u) - \v f^{wc}(\v x',\v u)]dt\|}} \cgr{+\max\{0,\phi(\v x')-\gamma dt\}- \epsilon} \nonumber\\
    \leq & \underline{k_{\phi}\cbl{\|\v x - \v x'\|} + k_{\phi}\cor{\|\v f^{wc}(\v x,\v u) - \v f^{wc}(\v x',\v u)\|}dt} + \max\{0,\phi(\v x')-\gamma dt\}- \epsilon \nonumber\\
    \leq& k_{\phi} \cbl{\delta} + k_{\phi} \underline{\|\v f^{wc}(\v x,\v u) - \v f(\v x,\v u)- \v f^{wc}(\v x',\v u) + \v f(\v x',\v u) + \v f(\v x,\v u)-\v f(\v x',\v u)}\| dt \notag\\ & + \max\{0,\phi(\v x)-\gamma dt\} +\cor{\max\{0,\phi(\v x')-\gamma dt\} - \max\{0,\phi(\v x)-\gamma dt\}}- \epsilon \nonumber\\
    \leq& k_{\phi} \delta + k_{\phi}( \underline{\cgr{\|\v f^{wc}(\v x,\v u) - \v f(\v x,\v u)\|} + \cgr{\| \v f(\v x',\v u)- \v f^{wc}(\v x',\v u)\|} + \cor{\|\v f(\v x,\v u)-\v f(\v x',\v u)\|}}) dt \notag\\ & + \max\{0,\phi(\v x)-\gamma dt\} +\cor{\|\phi(\v x') - \phi(\v x)\|}- \epsilon \nonumber\\
    \leq& k_{\phi} \delta + k_{\phi}( \cgr{\Delta f} + \cgr{\Delta f} + \cor{k_f\delta}) dt  + \max\{0,\phi(\v x)-\gamma dt\} +\cor{k_\phi \delta}- \epsilon \nonumber\\
   =  &\max\{0,\phi(\v x)-\gamma dt\}, 
\end{align}
which concludes the proof.
\end{proof}

\subsection{Linearization of the Safety Constraint with a Linear Taylor Layer}
Although the persistent feasibility is guaranteed by the worst-case safety index  $\phi$, we still need to address the nonlinearity of the safety constraint \eqref{eq:worst_case_safety}, which is equivalent to 
\begin{align}
\label{eq:constraint_equ}
    \phi(\v x_{k+1}^{wc}) := \max_{\v f(\v x_k, \v u_k)\in[ \v{\underline f}_{B}(\v x_k, \v u_k) ,\v{\bar f}_{B}(\v x_k, \v u_k) ]}\phi(\v x_k+ \v f(\v x_k, \v u_k) dt)  \leq \max\{0, \phi(\v x_{k}) - \gamma dt\}.
\end{align}
We apply the first-order Taylor expansion with Lagrange Mean Value Theorem for $\phi(\v x_{k+1})$ at the point $\v x_k$ for $\v x_{k+1} = \v x_k+ \v f(\v x_k, \v u_k) dt$ and obtain
\begin{align}
    \phi(\v x_{k+1}) = \underbrace{\phi(\v x_k) + \nabla_{\v x}^\top\phi(\v x_k)\v f(\v x_k, \v u_k)dt}_{\phi_{\v{f}}(\v x_k, \v u_k)} + \underbrace{\frac{1}{2}\v f(\v x_k, \v u_k)^\top \nabla_{\v x}^2\phi(\v x')\v f(\v x_k, \v u_k)(dt)^2}_{R^{\v x_{k}}(\v x')},
\end{align}
which consists of the first-order Taylor polynomial $\phi_{\v{f}}(\v x_k, \v u_k)$ and the Lagrange remainder term $R^{\v x_{k}}(\v x')$ with $\v x'\in[\v x_k, \v x_{k+1}]$. Then we formulate the first-order Taylor approximation $\phi_{\v{f}}(\v x_k, \v u_k): \real^{m_x} \times \real^{m_u} \mapsto \real$ as the composite function of the neural network $\v f(\v x_k, \v u_k): \real^{m_x} \times \real^{m_u} \mapsto \real^{m_x}$ and an additional linear mapping $\phi^{\v x_k}$ (called the linear Taylor layer) with weight $\nabla_{\v x}^\top\phi(\v x_k)dt\in \real^{1\times m_x}$ and bias $\phi(\v x_k)\in\real$, $\ie \phi_{\v f}=\phi^{\v x_k}\circ\v f, \phi^{\v x_k} (\v f) = \nabla_{\v x}^\top\phi(\v x_k)dt \v f + \phi(\v x_k)$. 

Similar to the computation of $\v{\underline f}_{B}(\v x, \v u)\leq \v f(\v x, \v u)\leq\v{\bar f}_{B}(\v x, \v u)$ in Section \ref{sec:BPO}, with BPO for each ReLU activation at each layer, the first-order Taylor approximation $\phi_{\v{f}} = \phi^{\v x_k}\circ\v{f}_n \circ \v{f}_{n-1} \circ \cdots \circ \v{f}_1$ can be relaxed to be $\underline{\phi}_{\v{f}_B}(\v x, \v u)\leq \phi_{\v{f}}(\v x, \v u)\leq\bar{\phi}_{\v{f}_B}(\v x, \v u)$ given $\v x, \v u$. For the  Lagrange remainder term $R^{\v x_{k}}(\v x')$ with $\v x'\in[\v x_k, \v x_{k+1}]$, we show that it can be bounded by $\frac{1}{2}M_f^2 M_\phi (dt)^2$ through Proposition \ref{prop: remainder}, while  $R^{\v x_{k}}(\v x')$ is usually neglected in the previous work \citep{wei2022safe}. 
\begin{proposition}\label{prop: remainder}
If the $\ell_2$ operator norm of  the Hessian matrix $\nabla_{\v x}^2\phi(\v x)$ is bounded by $M_\phi$ for any $\v x\in[\v x_k, \v x_{k+1}]$ and the Euclidean norm of $\v f(\v x, \v u)$ is bounded by $M_f$,  it holds that 
\begin{align}
\label{eq:bound_r}
    |R^{\v x_{k}}(\v x)|\leq \frac{1}{2}M_f^2 M_\phi (dt)^2, \quad \forall \v x\in[\v x_k, \v x_{k+1}]
\end{align}

\end{proposition}
\begin{proof}
    For $\v x\in[\v x_k, \v x_{k+1}]$, we have $\|\nabla_{\v x}^2\phi(\v x)\|_{op}\leq M_\phi, \|\v f(\v x, \v u)\|_2\leq M_f$, where \textit{op} indicates the operator norm. Therefore, we show Eq. \eqref{eq:bound_r} by using the \cbr{operator norm} and \uwave{Cauchy–Schwarz inequality} as below:   
\begin{align}
|R^{\v x_{k}}(\v x)| = & \frac{(dt)^2}{2} \cbr{\|\v f(\v x_k, \v u_k)^\top \nabla_{\v x}^2\phi(\v x')\v f(\v x_k, \v u_k)\|_2} \nonumber 
\leq\frac{(dt)^2}{2} \cbr{\|\v f(\v x_k, \v u_k)^\top\|_2 \|\nabla_{\v x}^2\phi(\v x')\v f(\v x_k, \v u_k)\|_{op}} \nonumber\\
\leq & \frac{(dt)^2}{2} \|\v f(\v x_k, \v u_k)^\top\|_2 \max_{\|\v f\|_2=1}\uwave{\|\v f^\top\nabla_{\v x}^2\phi(\v x')\v f(\v x_k, \v u_k)\|_{2}} \nonumber\\
\leq & \frac{(dt)^2}{2} \|\v f(\v x_k, \v u_k)^\top\|_2 \max_{\|\v f\|_2=1}\uwave{\|\v f\|_2\cbr{\|\nabla_{\v x}^2\phi(\v x')\v f(\v x_k, \v u_k)\|_{2}}} \nonumber\\
\leq & \frac{(dt)^2}{2} \|\v f(\v x_k, \v u_k)^\top\|_2 \cbr{\|\nabla_{\v x}^2\phi(\v x')\|_{op}\|\v f(\v x_k, \v u_k)\|_{2}} \nonumber
\leq \frac{1}{2}M_f^2 M_\phi (dt)^2.
\end{align}
which concludes the proof.
\end{proof}
Therefore, the safety constraint in  \eqref{eq:constraint_equ} can be rewritten as the linear inequality $\bar{\phi}_{\v{f}_B}(\v x_k, \v u_k) + \frac{1}{2}M_f^2 M_\phi (dt)^2 \leq \max\{0, \phi(\v x_{k}) - \gamma dt\}$, where $\bar{\phi}_{\v{f}_B}(\v x_k, \v u_k)$ is the upper bound of $\phi_{\v f}(\v x_k, \v u_k)$ with BPO relaxation and $\frac{1}{2}M_f^2 M_\phi (dt)^2$ can be approximated as an optimizable parameter in Sec. \ref{sec:worstcase_SI}.

\subsection{Safe Control with BPO-Relaxed NNDM}
Based on the worst case safety index (that ensures persistent feasibility) and the linearization of the worst-case safety constraint,  we finally transform the original constrained optimization \eqref{eq: bond opt} into the following form:
\begin{subequations}\label{eq:final_opt}
\begin{align}
&  \min_{\v u_k,\v x_{k+1},\{\v z_i\}_{i=0}^{n}} \|\v x_{k+1} - \v x_{k+1}^r\|_p\\ 
\st & \v x_{k+1} = \v x_k + \v z_n dt, ~ \v z_0 = [\v x_k^\top, \v u_k^\top ]^\top, ~ \v u_k\in \UU,\\
  \label{eq:model_z} 
& z_{i j} \geq \hat{z}_{i j}, ~
    z_{i j} \geq 0, ~
    z_{i j} \leq g_K(\hat z_{ij}) \text{ in \eqref{eq:BPO}}, \\
& \hat z_{ij} = \mathbf{w}_{i j} \mathbf{z}_{i-1}+b_{i j}, ~ \forall i \in\{1, \ldots, n\}, ~ \forall j \in\left\{1, \ldots, k_{i}\right\},  \label{eq:bpo_z}
    \\
& \bar{\phi}_{\v{f}_B}(\v x_k, \v u_k)  \leq \max\{-\zeta, \phi(\v x_{k}) - \gamma dt-\zeta\},
 \label{eq:safety_constraint}
\end{align}
\end{subequations}
where $\zeta = \frac{1}{2}M_f^2 M_\phi (dt)^2$ is from  Proposition \ref{prop: remainder}. 
In this paper, we consider $K=1,2$ in Eq. \ref{eq:BPO} as the BPO degree to illustrate the proposed design. 
When $K=1$, the optimization problem \eqref{eq:final_opt} is either Linear Programming ($p=1$) or Quadratic Programming ($p=2$); When $K=2$, it is either Quadratically Constrained Linear Programming ($p=1$) or Quadratically Constrained Quadratic Programming ($p=2$).
Besides, we approximate the upper bound  of $ \phi_{\v{f}_B}(\v x_k, \v u_k)$ in \eqref{eq:safety_constraint} by 
sampling and 
 explore the use of the existing solvers, including  CPLEX, Gurobi, and Ipopt, to solve the obtained optimization problems.

\section{Experiment}
\label{sec:experiment}
In the experiment, we aim to answer the following questions: how scalable is
the proposed BOND
compared to the MIP-based baseline \citep{wei2022safe} considering different sizes of models and tasks? How is the performance influenced by different optimization solvers and the tightness of BPO relaxation?
 We answer the first question in Section \ref{sec:collision}
 through the comparison of different dynamic models for collision avoidance and safe following of the unicycle, followed by the validation of the effectiveness of the worst-case safety constraint. Section \ref{sec:ablation} shows the influence of several key factors for the second question.

\subsection{Experimental Setup}
\paragraph{Environment and dynamics.}
To be consistent with \cite{wei2022safe}, the experiment is based on the neural network dynamic models for a second-order unicycle in a 2D setting. The 4D states $\XX\subset \real^4$ are the 2D positions, velocity and heading angle, and the 2D control inputs $\UU\subset \real^2$ are the acceleration and angular velocity. The current states and control inputs are also the inputs of neural networks, and the outputs of the neural networks are the 2D velocity, acceleration, and angular velocity as the derivatives of the states. The states and inputs are bounded as $B\subset \XX:  [-10,10]\times[-10,10]\times[-2,2]\times[-\pi,\pi]$ and $\UU: [-4,4]\times[-\pi,\pi]$. Collision avoidance and safe following are used for evaluation with different safety constraints, where the unicycle is supposed to be at least 0.5 away from the obstacle for collision avoidance and be within 1 and 2 away from the moving target for safe following.
 The neural networks have fully-connected layers with the ReLU activation, with different depths ($d=2,3,4$) and widths ($w=50, 100,200$) and are denoted as FC$d$-$w$, \eg FC3-100 means a model of 3-layer with 100 neurons per layer. 
To verify our small models ($<1000$ neurons), MIP works the best according to $\alpha,\beta$-CROWN \citep{wang2021beta}.

\paragraph{Optimization and evaluation metrics.} 
To solve the real-time optimization in \eqref{eq:final_opt}, the reference states are generated through one-tenth interpolation between the current state and the goal as real-time planning.
We solve the optimization using CPLEX, Gurobi and Ipopt 
with linear or quadratic objectives ($p=1,2$, respectively) of the tracking error term in \eqref{eq:safe_control_minimax} for both the baseline and our BPO relaxation of the degree of 1 and 2 ($K=1,2$). The pre-activation bounds are computed using ConvDual \citep{wong2018provable} and interval
arithmetic (IA), where the former is much tighter \citep{liu2021algorithms,gowal2018effectiveness}.  The default setting is with ConvDual pre-activation bounds under CPLEX solver, for both baseline and our 1-degree BPO, while our 2-degree BPO is with Ipopt solver as default due to nonconvex quadratic constraints. More results regarding these factors can be found in Section \ref{sec:ablation}. The evaluation metrics are \textit{prediction time per step} and \textit{prediction error per step}, where the latter is between the optimized and the executed states with corresponding norms in the optimization objective. The mean of each metric is calculated for 10  trajectories with random initial states under each setting, where the step number per trajectory is around 100.  The code is available at \url{https://github.com/intelligent-control-lab/BOND}.  



\subsection{Performance Comparison with Baseline}
\label{sec:collision}


\begin{figure}
    \centering
    \subfigure[]{\includegraphics[width=0.26\textwidth]{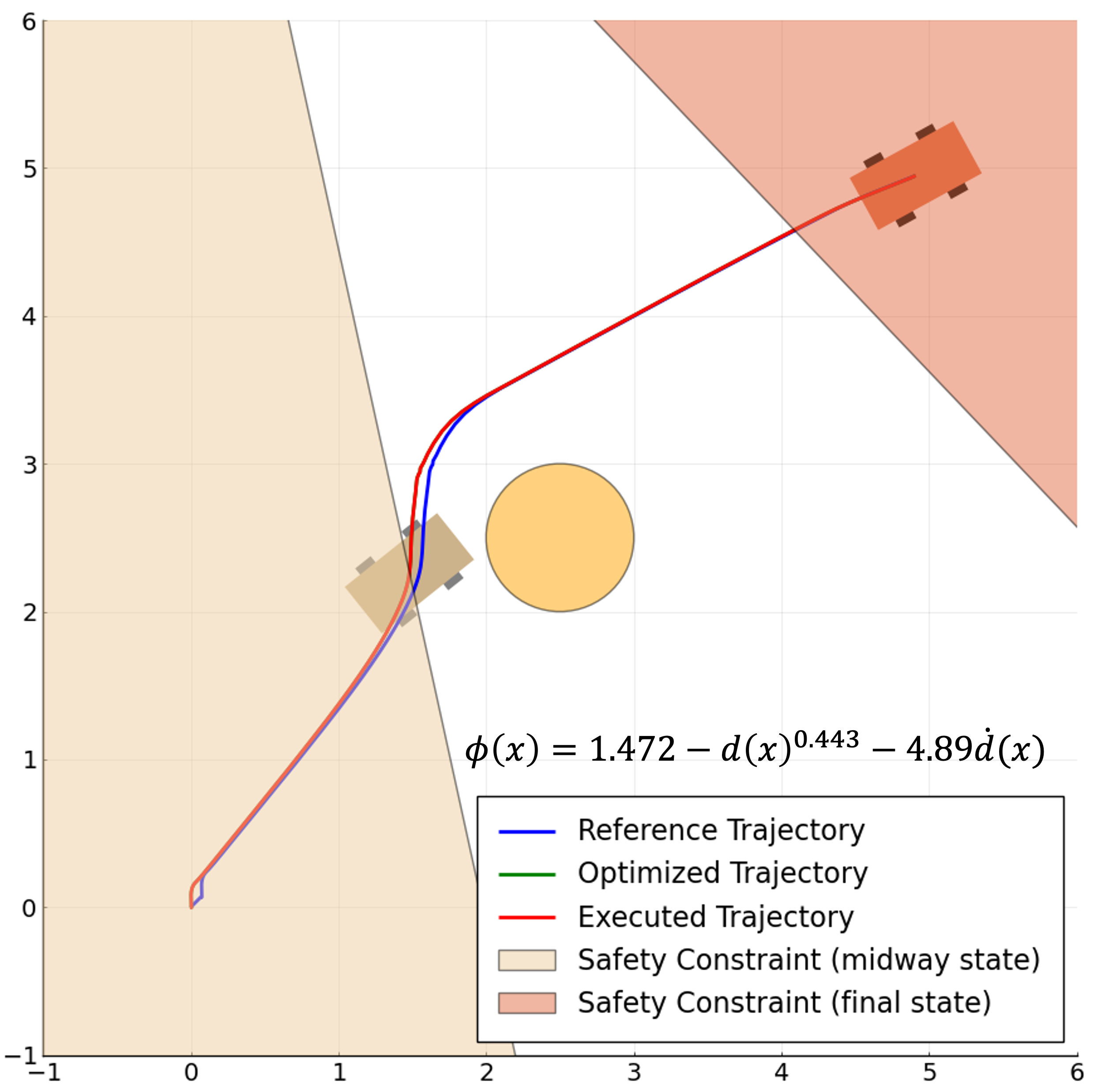}
    } 
    \subfigure[]{\includegraphics[width=0.26\textwidth]{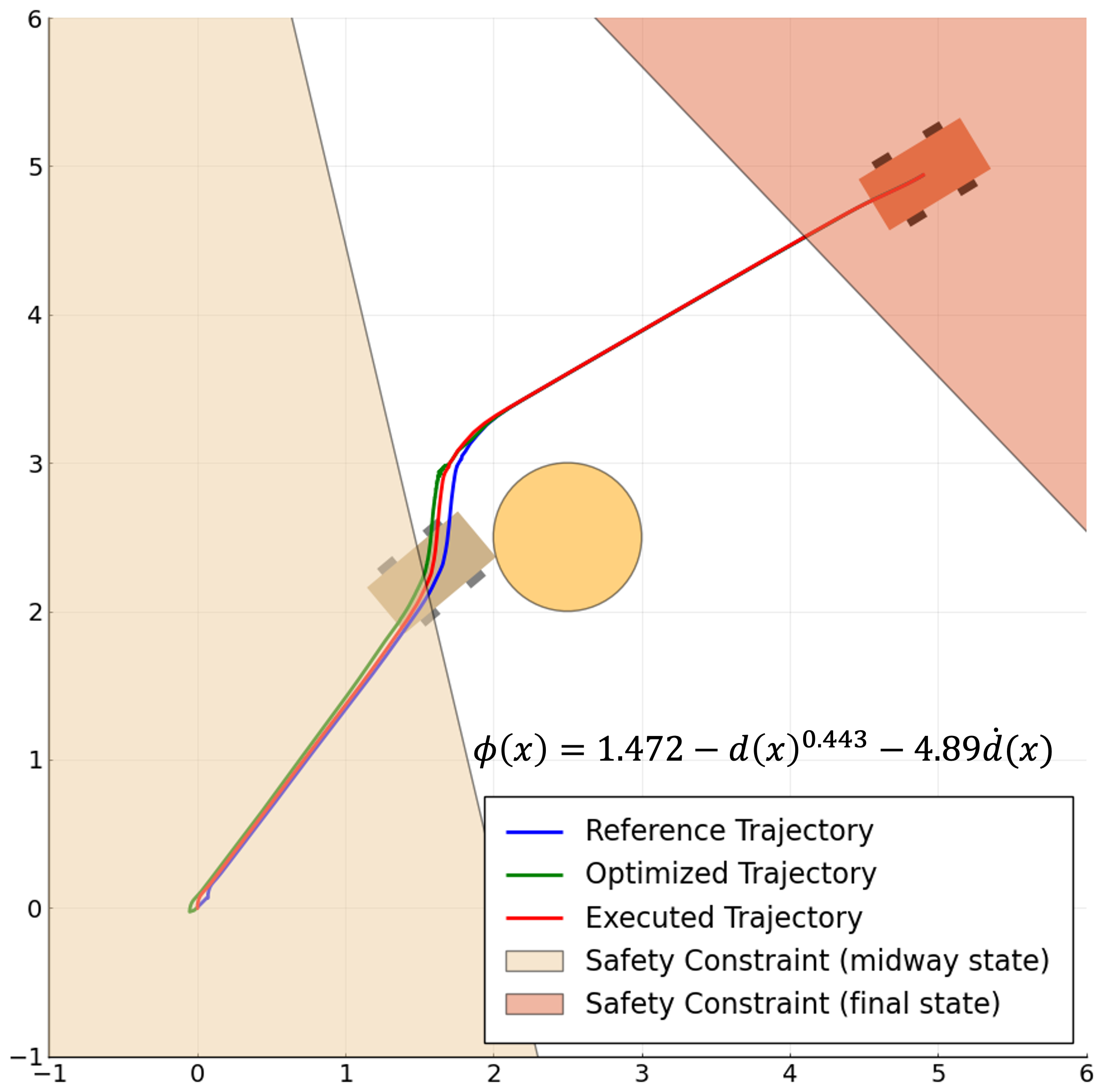}} 
    \subfigure[]{\includegraphics[width=0.26\textwidth]{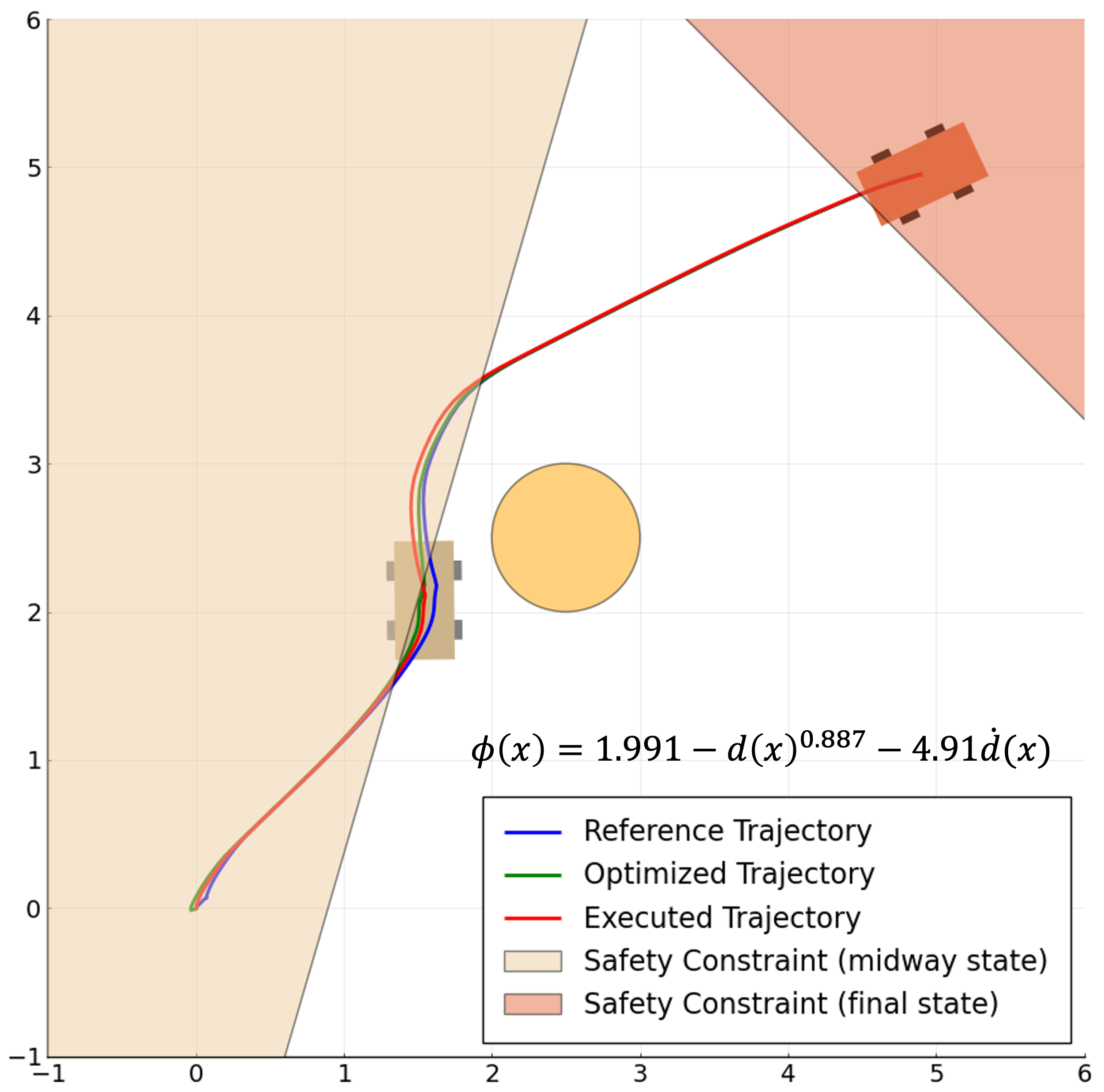}}
    \vspace{-3mm}
    \caption{Collision avoidance  with (a)  MIP-based NNDM with plain safety constraint; (b) BOND with plain safety constraint; (c) BOND with worst-case safety constraint. The safety constraint is visualized as the half-space of state space based on  \eqref{eq:safety_constraint}. $\phi(x)$ is the corresponding safety index and $d(x)$ is the distance between the unicycle and obstacle.}
    \vspace{-3mm}
    \label{fig:wc_SIS}
\end{figure}
 \begin{table}[]
     \centering
     \resizebox{0.80\textwidth}{!}{
     \begin{tabular}{cccccccc}
     \toprule
\multicolumn{2}{c}{\multirow{2}{*}{\begin{tabular}[c]{@{}c@{}}Collision \\ Avoidance\end{tabular}}} & \multicolumn{2}{c}{FC2-100}      & \multicolumn{2}{c}{FC3-100}      & \multicolumn{2}{c}{FC4-100}      \\ \cmidrule(r){3-4} \cmidrule(r){5-6} \cmidrule(r){7-8}
\multicolumn{2}{c}{} &  Error & Time (s) &  Error & Time (s) &  Error & Time (s) \\ \midrule
\multirow{3}{*}{\begin{tabular}[c]{@{}c@{}}Linear \\ Objective\end{tabular}} & MIND-SIS & $\mathbf{2.31e^{-9}}$ 
& 0.0582  &   $\mathbf{1.56e^{-9}}$   & \underline{0.255}    &   $\mathbf{1.42e^{-9}}$    & 151    \\
  & Ours (BPO-1)     &0.106 &\textbf{0.0031}      &0.0943&  \textbf{0.0080}  & 0.0911&\textbf{0.0279}    \\ & Ours (BPO-2)     & \underline{0.0668}&  \underline{0.0296}    & \underline{0.0298}  & 0.716    & \underline{0.0757} &  \underline{2.76}    \\\cmidrule(r){1-2}
\multirow{3}{*}{\begin{tabular}[c]{@{}c@{}}Quadratic\\ Objective\end{tabular}}  & MIND-SIS &   $\mathbf{1.78e^{-9}}$      & 0.0576       & $\mathbf{1.31e^{-9}}$ &  \underline{0.338}    & $\mathbf{1.29e^{-9}}$ &   226   \\
  & Ours (BPO-1)     & 0.0435 & \textbf{0.0085}     & 0.0338  &  \textbf{0.0222}      & 0.0401& \textbf{ 0.211}   \\ & Ours (BPO-2)     & \underline{0.0263}&   \underline{0.0191}   & \underline{0.0191}  & 0.577      & \underline{0.0273}& \underline{2.05}     \\\bottomrule
\end{tabular}
}
\vspace{-2mm}
\caption{Comparison of the baseline and ours under different model complexity and optimization objective norms ($p=1,2$)  for collision avoidance. Notations: \textbf{best} and \underline{second best} results.
     }
     \label{tab:collision}
     \vspace{-6mm}
 \end{table}

\paragraph{Significance of the worst-case safety constraint for BPO-relaxed NNDM.} As shown in Fig. \ref{fig:wc_SIS}(a), the MIP-based baseline MIND-SIS \citep{wei2022safe} works well under the plain safety constraint of $\XX_S$ in \cite{wei2022safe}, while the plain safety constraint results in prediction error between the optimized and executed states under BOND, causing collision as (b) shows. However, with the proposed worst-case safety constraint \eqref{eq:BPO_safe} of $\XX_{BPO}$, collision avoidance under BOND is guaranteed in (c) even if the prediction error still exists between the optimized and executed states. This validates the significance of the more conservative worst-case safety constraint \eqref{eq:BPO_safe} for BPO-relaxed NNDM. The offline time for synthesizing the plain safety index is  4.57h for (a) and (b), while the time for our worst-case one is 19.4h due to higher computation complexity.



\paragraph{Performance comparison of prediction error and computation time.}
Table \ref{tab:collision} and Table \ref{tab:following} present the results of baseline MIND-SIS \citep{wei2022safe} and ours with BPO degrees of 1 and 2 for collision avoidance and safe following. It can be seen that under all depths of models, our BPO relaxation results in 10-100 times less computation time per step compared to the baseline, although their prediction errors are larger than those of the baseline as ground truth. Across all the models, BPO-2 has smaller prediction errors but slower computation than BPO-1 due to tighter but non-convex quadratic relaxation in \eqref{eq:BPO} when $K=2$. As the models go deeper, we can see our prediction time does not drastically increase as MIND-SIS does, showing our method scales better.

 \begin{table}[]
     \centering
     \resizebox{0.80\textwidth}{!}{
  \begin{tabular}{cccccccc}
     \toprule
\multicolumn{2}{c}{\multirow{2}{*}{\begin{tabular}[c]{@{}c@{}}Safe \\ Following\end{tabular}}} & \multicolumn{2}{c}{FC2-100}      & \multicolumn{2}{c}{FC3-100}      & \multicolumn{2}{c}{FC4-100}      \\ \cmidrule(r){3-4} \cmidrule(r){5-6} \cmidrule(r){7-8}
\multicolumn{2}{c}{} & Error & Time (s) & Error & Time (s) & Error & Time (s) \\ \midrule
\multirow{3}{*}{\begin{tabular}[c]{@{}c@{}}Linear \\ Objective\end{tabular}} & MIND-SIS &   $\mathbf{2.86e^{-9}}$  & 0.0387    & $\mathbf{1.55e^{-9}}$& \underline{0.232}  &  $\mathbf{9.34e^{-10}}$ & 40.4    \\
  & Ours (BPO-1)     &0.116  & \textbf{0.0034}     & 0.0886 &  \textbf{0.0072}    &0.112 &  \textbf{0.444}    \\ & Ours (BPO-2)     &\underline{0.0504} &  \underline{0.0360}    &\underline{0.0518}  &    0.579  &\underline{0.0794} &  \underline{2.40}    \\\cmidrule(r){1-2}
\multirow{3}{*}{\begin{tabular}[c]{@{}c@{}}Quadratic\\ Objective\end{tabular}}  & MIND-SIS & $\mathbf{1.55e^{-9}}$ &  0.0517    &   $\mathbf{1.12e^{-9}}$     & \underline{0.267}      &$\mathbf{1.08e^{-9}}$& 52.8     \\
  & Ours (BPO-1)     &  0.0243&\textbf{0.0071}      &0.0417 & \textbf{0.0223}     & 0.0498 &\textbf{0.455}   \\ & Ours (BPO-2)     & \underline{0.0241} &  \underline{0.0283}    &\underline{0.0249}  &  0.503    &  \underline{0.0477}&  \underline{2.08}    \\\bottomrule
\end{tabular}
}
\vspace{-1mm}
\caption{Comparison of the baseline and ours under different model complexity and optimization objective norms ($p=1,2$) for safe following. Notations: \textbf{best} and \underline{second best} results.
     }
     \label{tab:following}
     \vspace{-2mm}
 \end{table}
\begin{table}[]
    \centering
    \resizebox{0.80\textwidth}{!}{
    \begin{tabular}{cccccccc}
\toprule
\multicolumn{2}{c}{\multirow{2}{*}{\begin{tabular}[c]{@{}c@{}}Different solvers\\for two tasks\end{tabular}}} & \multicolumn{2}{c}{MIND-SIS}     & \multicolumn{2}{c}{Ours (BPO-1)}& \multicolumn{2}{c}{Ours (BPO-2)}\\ \cmidrule(r){3-4} \cmidrule(r){5-6} \cmidrule(r){7-8} 
\multicolumn{2}{c}{}   & Error & Time (s) & Error & Time (s) & Error & Time (s) \\ \midrule
\multicolumn{1}{c}{\multirow{3}{*}{\begin{tabular}[c]{@{}c@{}}Collision\\ Avoidance\end{tabular}}}  & CPLEX   &  $\mathbf{1.31e^{-9}}$ &  \textbf{0.338}      & 0.0338 &  \textbf{0.0222}   & ---    &  ---    \\ 
 & Gurobi  &  $2.02\text{e}^{-9}$   &  0.516    & \textbf{0.0327}  &  0.0572    &   0.0396   &   529   \\ 
 & Ipopt   & ---   &   ---   &   0.0332  &   0.338   & \textbf{0.0191} & \textbf{0.577}      \\ \cmidrule{1-2}
\multicolumn{1}{c}{\multirow{3}{*}{\begin{tabular}[c]{@{}c@{}}Safe\\ Following\end{tabular}}}  & CPLEX   & $\mathbf{1.12e^{-9}}$     & \textbf{0.267}       & \textbf{0.0417} & \textbf{0.0223}      & ---    &   ---   \\ 
 & Gurobi  &   $1.25\text{e}^{-9}$  &  0.349    &  0.0418  &   0.0475   &  0.0495   &     333 \\ 
 & Ipopt   & ---   &   ---   &  0.0439   & 0.317     & \textbf{0.0249} &  \textbf{0.503}      \\ \bottomrule
\end{tabular}
}
\vspace{-1mm}
    \caption{Comparison of performance with different solvers using FC3-100 and quadratic objective for both baseline and ours. The best results among different solvers are in \textbf{bold} and ``---'' indicates infeasibility.}
    \vspace{-5mm}
    \label{tab:BPO_degree_solver}
\end{table}

\subsection{Ablation Study}
\label{sec:ablation}

\paragraph{Influence of optimization solvers.} Table \ref{tab:BPO_degree_solver} shows how commonly used solvers affect the tracking performance. We can find that CPLEX is usually the fastest for MIND-SIS and our BPO-1, but it cannot solve BPO-2 with non-convex quadratic constraints. Gurobi generally applies to solving different problems but suffers from longer computation time, especially for BPO-2. As a nonlinear optimizer, Ipopt performs satisfactory results for BPO-2 with local convergence at risk of unsoundness, and it is much slower for BPO-1 and cannot be used for the MIP-based baseline. 

\begin{table}[]
    \centering
    \resizebox{0.80\textwidth}{!}{
\begin{tabular}{cccccccc}
\toprule
\multicolumn{2}{c}{\multirow{2}{*}{\begin{tabular}[c]{@{}c@{}}Different pre- \\ activation bounds\end{tabular}}} & \multicolumn{2}{c}{MIND-SIS}     & \multicolumn{2}{c}{Ours (BPO-1)} & \multicolumn{2}{c}{Ours (BPO-2)} \\ \cmidrule(r){3-4} \cmidrule(r){5-6} \cmidrule(r){7-8} 
\multicolumn{2}{c}{}& Error & Time (s) & Error & Time (s) & Error & Time (s) \\ \midrule
\multicolumn{1}{c}{\multirow{2}{*}{\begin{tabular}[c]{@{}c@{}}Collision \\ Avoidance\end{tabular}}}& IA     &   $1.34\text{e}^{-9}$ &   0.511   &  0.0385   & 0.0260     &  0.0254  & 0.812     \\  
& ConvDual&  $\mathbf{1.31e^{-9}}$&  \textbf{0.338}      & \textbf{0.0338} &  \textbf{0.0222}      &\textbf{0.0191}  & \textbf{0.577}      \\ \cmidrule{1-2}
\multicolumn{1}{c}{\multirow{2}{*}{\begin{tabular}[c]{@{}c@{}}Safe \\ Following\end{tabular}}}& IA     &    $1.17\text{e}^{-9}$  &  0.365   &  0.0554   &   0.0265   &  0.0510   &  1.12    \\ 
& ConvDual&        $\mathbf{1.12e^{-9}}$       & \textbf{0.267} & \textbf{0.0417} & \textbf{0.0223}      &\textbf{0.0249}  &  \textbf{0.503}      \\ \bottomrule
\end{tabular}
}
\vspace{-1mm}
    \caption{Comparison of performance with different pre-activation bounds for both baseline and ours  using  FC3-100 and quadratic objective. The better results between IA and ConvDual are in \textbf{bold}.}
    \vspace{-2mm}
    \label{tab:pre-activation}
\end{table}
\begin{figure}
    \centering
    \subfigure[]{\includegraphics[width=0.32\textwidth]{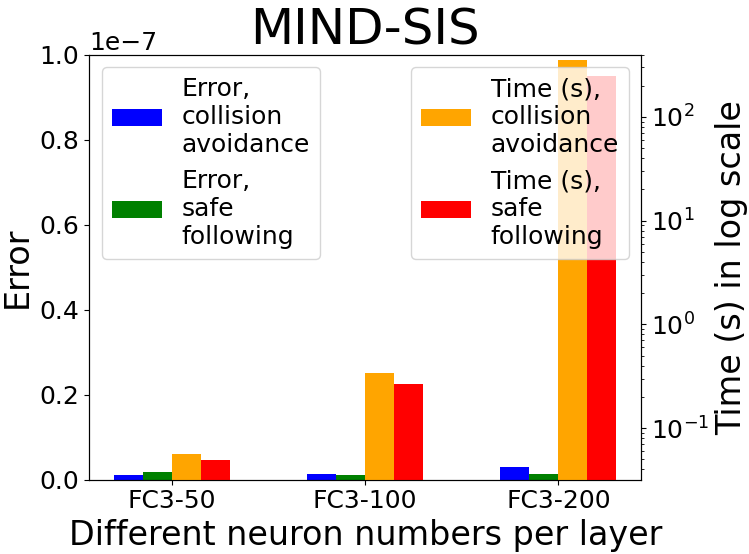}
    } 
    \subfigure[]{\includegraphics[width=0.32\textwidth]{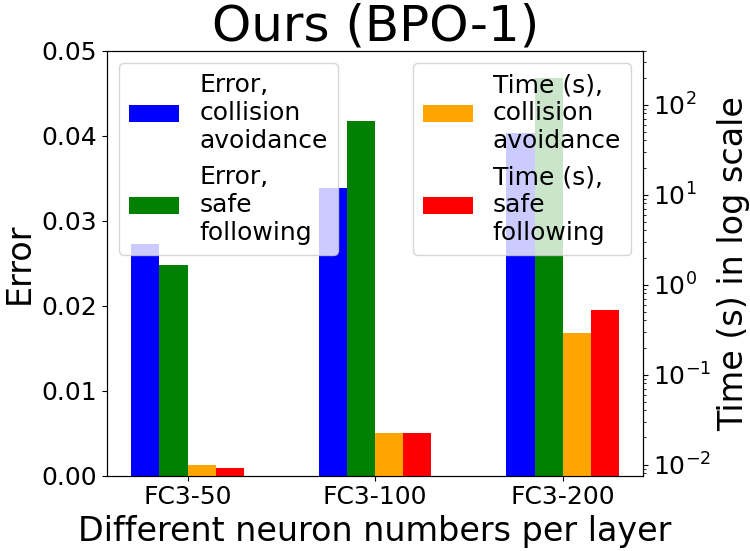}} 
    \subfigure[]{\includegraphics[width=0.32\textwidth]{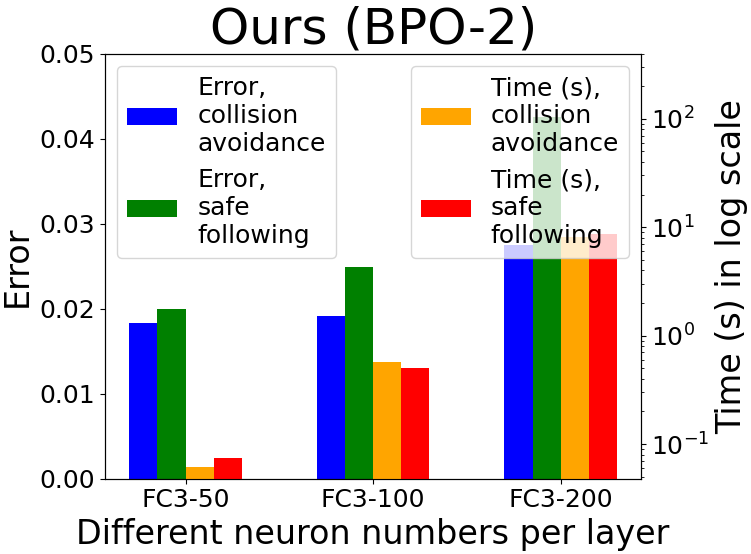}}
    \vspace{-3mm}
    \caption{Comparison of performance with different neuron numbers per layer for both baseline and ours using quadratic objective. (a)  MIND-SIS. (b) Ours (BPO-1). (c) Ours (BPO-2).}
    \vspace{-5mm}
    \label{fig:neuron_num}
\end{figure}

\paragraph{Influence of pre-activation bounds.} Since the MIP-based baseline and ours both greatly rely on the pre-activation bounds of  ReLU activation layers, we compare the results of different pre-activation bounds, interval arithmetic (IA) \citep{liu2021algorithms,gowal2018effectiveness} and ConvDual \citep{wong2018provable}, as shown in Table \ref{tab:pre-activation}. It can be seen that IA causes longer prediction time and larger errors per step because of its poor tightness and large search space. 

\paragraph{Influence of neuron numbers per layer.} From Fig. \ref{fig:neuron_num}, it can be seen that the time consumption of MIND-SIS exponentially explodes when layer width increases, while our BPO-based ones maintain a relatively linearly increased computation time, validating the remarkable scalability of our methods. Different from Table \ref{tab:collision} and Table \ref{tab:following}, the errors increase as the neuron number per layer goes up under BPO relaxation, implying that the relaxation becomes looser with more neurons per layer.


\section{Conclusion}
\label{sec:conclusion}
In this work,  we introduce Bernstein over-approximated neural dynamics (BOND) with  Bernstein polynomial over-approximation (BPO) of ReLU activation layers in NNDMs to speed up the optimization of safe tracking. To ensure the persistent feasibility of safety set under the approximation error of BOND, the worst-case safety index is synthesized offline to satisfy the safety constraint for the most unsafe potential predicted states of BOND. 
Comprehensive experiments validate the time efficiency and scalability of BOND. Our main limitation lies in the trade-off between optimality and conservativeness due to the worst-case safety constraint. Besides, the model mismatch has not been considered in our setting.
Future directions can be exploring the robustness of BOND in more real-world robot settings to ensure safety under out-of-distribution model mismatch.

\clearpage
\acks{This work is in part supported by  the National Science Foundation under Grant No. 2144489. Any opinions, findings, and conclusions or recommendations expressed in this material are those of the authors and do not necessarily reflect the views of the National Science Foundation. Jianglin Lan is supported by a Leverhulme Trust Early Career Fellowship under Award ECF-2021-517. We also would like to thank Tianhao Wei from CMU for the support in the experiments.}

\bibliography{l4dc2024.bib}

\end{document}